\RequirePackage{fix-cm}
\documentclass[twocolumn]{svjour3}          
\smartqed  
\usepackage{graphicx}
\usepackage{amsmath,amssymb} 
\usepackage{color}
\usepackage{bm}
\usepackage{tabularx}
\usepackage{url}
\usepackage{natbib}
\usepackage{apalike}
\usepackage{footnote}
\makesavenoteenv{tabular}

\newcolumntype{Y}{>{\centering\arraybackslash}X}

\makeatletter
\def\thickhline{%
    \noalign{\ifnum0=`}\fi\hrule \@height \thickarrayrulewidth \futurelet
    \reserved@a\@xthickhline}
\def\@xthickhline{\ifx\reserved@a\thickhline
    \vskip\doublerulesep
    \vskip-\thickarrayrulewidth
    \fi
    \ifnum0=`{\fi}}
\makeatother
\newlength{\thickarrayrulewidth}
\begin{document}

\title{CornerNet: Detecting Objects as Paired Keypoints
}


\author{Hei Law \and Jia Deng}


\institute{H. Law \at
           Princeton University, Princeton, NJ, USA \\
           \email{heilaw@cs.princeton.edu}
           \and
           J. Deng \at
           Princeton Universtiy, Princeton, NJ, USA
}


\maketitle

\begin{abstract}
    We propose CornerNet, a new approach to object detection where we
    detect an object bounding box as a pair of keypoints, the top-left
    corner and the bottom-right corner, using a single convolution neural
    network. By detecting objects as paired keypoints, we eliminate the
    need for designing a set of anchor boxes commonly used in prior
    single-stage detectors. In addition to our novel formulation, we
    introduce corner pooling, a new type of pooling layer that helps the
    network better localize corners.  Experiments show that CornerNet
    achieves a 42.2\% AP on MS COCO, outperforming all existing one-stage
    detectors. 
\end{abstract}

\section{Introduction} 
Object detectors based on convolutional neural networks
(ConvNets)~\citep{krizhevsky2012imagenet,simonyan2014very,he2016deep} have
achieved state-of-the-art results on various challenging
benchmarks~\citep{lin2014microsoft,deng2009imagenet,everingham2015pascal}.
A common component of state-of-the-art approaches is anchor
boxes~\citep{ren2015faster,liu2016ssd}, which are boxes of various sizes and
aspect ratios that serve as detection candidates. Anchor boxes are
extensively used in one-stage
detectors~\citep{liu2016ssd,fu2017dssd,redmon2016yolo9000,lin2017focal},
which can achieve results highly competitive with two-stage
detectors~\citep{ren2015faster,girshick2014rich,girshick2015fast,he2017mask}
while being more efficient. One-stage detectors place anchor boxes densely
over an image and generate final box predictions by scoring anchor boxes
and refining their coordinates through regression. 

But the use of anchor boxes has two drawbacks. First, we typically need a
very large set of anchor boxes, e.g. more than 40k in
DSSD~\citep{fu2017dssd} and more than 100k in
RetinaNet~\citep{lin2017focal}. This is because the detector is trained to
classify whether each anchor box sufficiently overlaps with a ground truth
box, and a large number of anchor boxes is needed to ensure sufficient
overlap with most ground truth boxes. As a result, only a tiny fraction of
anchor boxes will overlap with ground truth; this creates a huge imbalance
between positive and negative anchor boxes and slows down
training~\citep{lin2017focal}.

Second, the use of anchor boxes introduces many hyperparameters and design
choices. These include how many boxes, what sizes, and what aspect ratios.
Such choices have largely been made via ad-hoc heuristics, and can become
even more complicated when combined with multiscale architectures where a
single network makes separate predictions at multiple resolutions, with
each scale using different features and its own set of anchor
boxes~\citep{liu2016ssd,fu2017dssd,lin2017focal}. 

\begin{figure*}
    \centering
    \resizebox{0.8\textwidth}{!}{\includegraphics{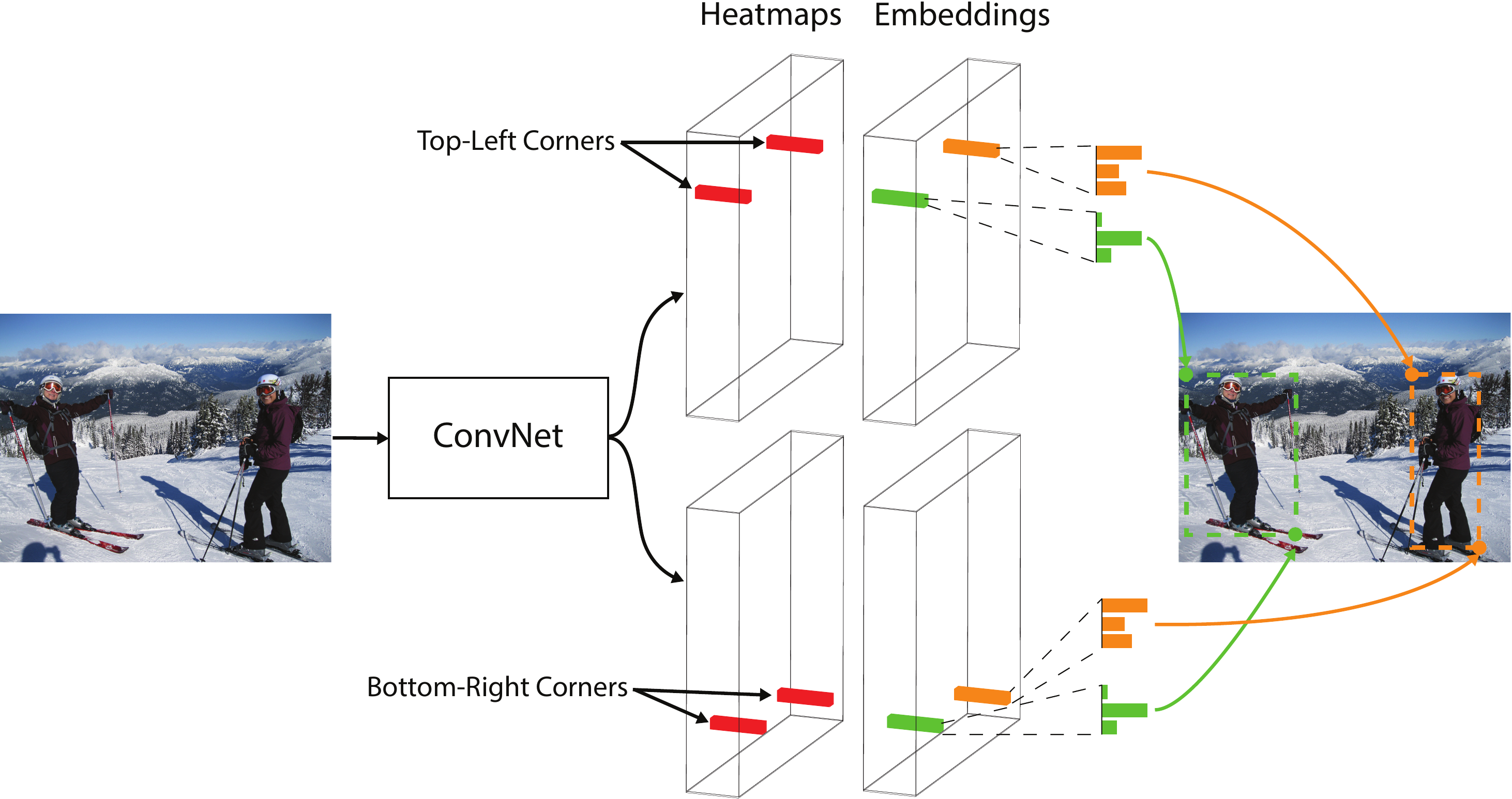}}
    \caption{We detect an object as a pair of bounding box corners grouped
    together. A convolutional network outputs a heatmap for all top-left
    corners, a heatmap for all bottom-right corners, and an embedding
    vector for each detected corner. The network is trained to predict
    similar embeddings for corners that belong to the same object.}
    \label{fig:approach}
\end{figure*}

In this paper we introduce CornerNet,  a new one-stage approach to object
detection that does away with anchor boxes. We detect an object as a pair
of keypoints---the top-left corner and bottom-right corner of the bounding
box. We use a single convolutional network to predict a heatmap for the
top-left corners of all instances of the same object category, a heatmap
for all bottom-right corners, and an embedding vector for each detected
corner. The embeddings serve to group a pair of corners that belong to the
same object---the network is trained to predict similar embeddings for
them. Our approach greatly simplifies the output of the network and
eliminates the need for designing anchor boxes. Our approach is inspired by
the associative embedding method proposed by \cite{newell2017associative},
who detect and group keypoints in the context of multiperson human-pose
estimation. Fig.~\ref{fig:approach} illustrates the overall pipeline of our
approach. 

\begin{figure*}
    \centering
    \resizebox{\textwidth}{!}{\includegraphics{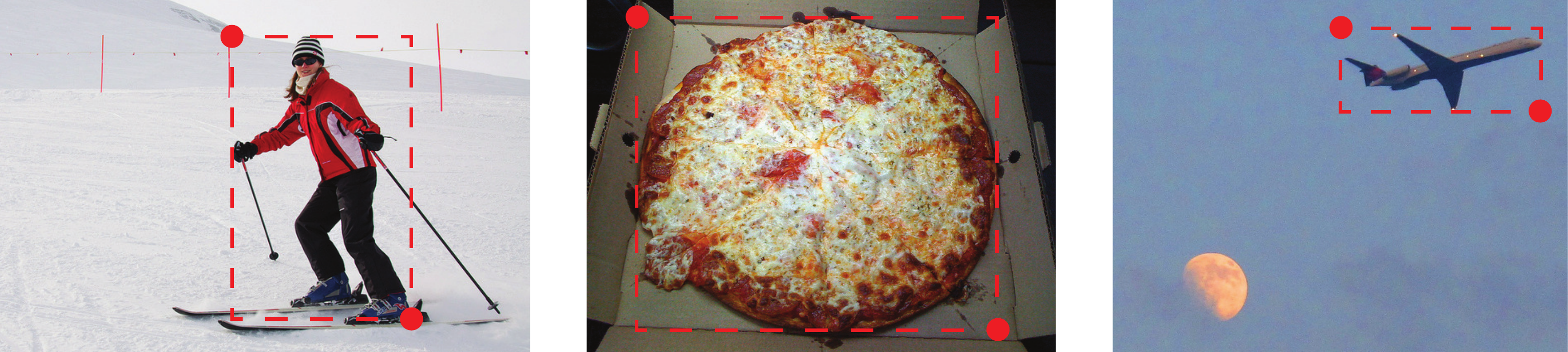}}
    \caption{Often there is no local evidence to determine the location of
    a bounding box corner. We address this issue by proposing a new type of
    pooling layer.}
    \label{fig:intro_local_evidence}
\end{figure*}

\begin{figure*}
    \centering
    \resizebox{0.6\textwidth}{!}{\includegraphics{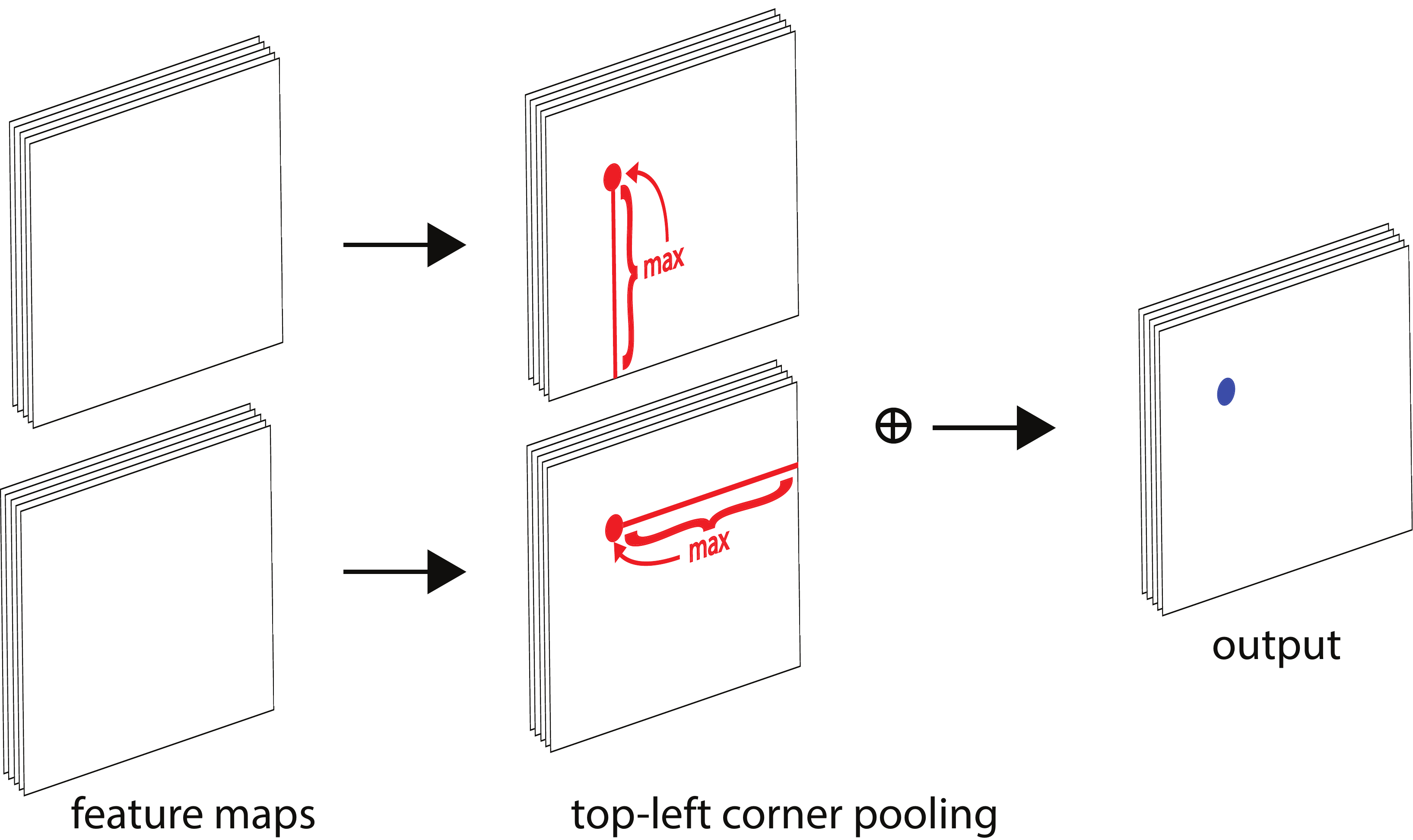}}
    \caption{Corner pooling: for each channel, we take the maximum values
    \textit{(red dots)} in two directions \textit{(red lines)}, each from a
    separate feature map, and add the two maximums together \textit{(blue
    dot)}.}
    \label{fig:intro_pooling}
\end{figure*}

Another novel component of CornerNet is \textit{corner pooling}, a new type
of pooling layer that helps a convolutional network better localize corners
of bounding boxes. A corner of a bounding box is often outside the
object---consider the case of a circle as well as the examples in
Fig.~\ref{fig:intro_local_evidence}. In such cases a corner cannot be
localized based on local evidence. Instead, to determine whether there is a
top-left corner at a pixel location, we need to look horizontally towards
the right for the topmost boundary of the object, and look vertically
towards the bottom for the leftmost boundary. This motivates our corner
pooling layer: it takes in two feature maps; at each pixel location it
max-pools all feature vectors to the right from the first feature map,
max-pools all feature vectors directly below from the second feature map,
and then adds the two pooled results together. An example is shown in
Fig.~\ref{fig:intro_pooling}.

We hypothesize two reasons why detecting corners would work better than
bounding box centers or proposals. First, the center of a box can be harder
to localize because it depends on all 4 sides of the object, whereas
locating a corner depends on 2 sides and is thus easier, and even more so
with corner pooling, which encodes some explicit prior knowledge about the
definition of corners. Second, corners provide a more efficient way of
densely discretizing the space of boxes: we just need $O(wh)$ corners to
represent $O(w^2h^2)$ possible anchor boxes.

We demonstrate the effectiveness of CornerNet on MS
COCO~\citep{lin2014microsoft}. CornerNet achieves a 42.2\% AP,
outperforming all existing one-stage detectors. In addition, through
ablation studies we show that corner pooling is critical to the superior
performance of CornerNet. Code is available at
\url{https://github.com/princeton-vl/CornerNet}.

\section{Related Works}
\subsection{Two-stage object detectors} Two-stage approach was first
introduced and popularized by R-CNN~\citep{girshick2014rich}. Two-stage
detectors generate a sparse set of regions of interest (RoIs) and classify
each of them by a network. R-CNN generates RoIs using a low level vision
algorithm~\citep{uijlings2013selective,zitnick2014edge}. Each region is
then extracted from the image and processed by a ConvNet independently,
which creates lots of redundant computations. Later,
SPP~\citep{he2014spatial} and Fast-RCNN~\citep{girshick2015fast} improve
R-CNN by designing a special pooling layer that pools each region from
feature maps instead. However, both still rely on separate proposal
algorithms and cannot be trained end-to-end.
Faster-RCNN~\citep{ren2015faster} does away low level proposal algorithms
by introducing a region proposal network (RPN), which generates proposals
from a set of pre-determined candidate boxes, usually known as anchor
boxes.  This not only makes the detectors more efficient but also allows
the detectors to be trained end-to-end. R-FCN~\citep{dai2016r} further
improves the efficiency of Faster-RCNN by replacing the fully connected
sub-detection network with a fully convolutional sub-detection network.
Other works focus on incorporating sub-category
information~\citep{xiang2016subcategory}, generating object proposals at
multiple scales with more contextual
information~\citep{bell2016inside,cai2016unified,shrivastava2016beyond,lin2016feature},
selecting better features~\citep{zhai2017feature}, improving
speed~\citep{li2017light}, cascade procedure~\citep{cai2017cascade} and
better training procedure~\citep{singh2017analysis}.

\subsection{One-stage object detectors} 
On the other hand, YOLO~\citep{redmon2016you} and SSD~\citep{liu2016ssd}
have popularized the one-stage approach, which removes the RoI pooling step
and detects objects in a single network. One-stage detectors are usually
more computationally efficient than two-stage detectors while maintaining
competitive performance on different challenging benchmarks. 

SSD places anchor boxes densely over feature maps from multiple scales,
directly classifies and refines each anchor box. YOLO predicts bounding box
coordinates directly from an image, and is later improved in
YOLO9000~\citep{redmon2016yolo9000} by switching to anchor boxes.
DSSD~\citep{fu2017dssd} and RON~\citep{kong2017ron} adopt networks similar
to the hourglass network~\citep{newell2016stacked}, enabling them to
combine low-level and high-level features via skip connections to predict
bounding boxes more accurately. However, these one-stage detectors are
still outperformed by the two-stage detectors until the introduction of
RetinaNet~\citep{lin2017focal}. In~\citep{lin2017focal}, the authors
suggest that the dense anchor boxes create a huge imbalance between
positive and negative anchor boxes during training. This imbalance causes
the training to be inefficient and hence the performance to be suboptimal.
They propose a new loss, Focal Loss, to dynamically adjust the weights of
each anchor box and show that their one-stage detector can outperform the
two-stage detectors. RefineDet~\citep{zhang2017single} proposes to filter
the anchor boxes to reduce the number of negative boxes, and to coarsely
adjust the anchor boxes. \\

DeNet~\citep{tychsen2017denet} is a two-stage detector which generates RoIs
without using anchor boxes. It first determines how likely each location
belongs to either the top-left, top-right, bottom-left or bottom-right
corner of a bounding box. It then generates RoIs by enumerating all
possible corner combinations, and follows the standard two-stage approach
to classify each RoI. Our approach is very different from DeNet. First,
DeNet does not identify if two corners are from the same objects and relies
on a sub-detection network to reject poor RoIs. In contrast, our approach
is a one-stage approach which detects and groups the corners using a single
ConvNet. Second, DeNet selects features at manually determined locations
relative to a region for classification, while our approach does not
require any feature selection step. Third, we introduce corner pooling, a
novel type of layer to enhance corner detection. 

Point Linking Network (PLN)~\citep{wang2017point} is an one-stage detector
without anchor boxes. It first predicts the locations of the four corners
and the center of a bounding box. Then, at each corner location, it
predicts how likely each pixel location in the image is the center.
Similarly, at the center location, it predicts how likely each pixel
location belongs to either the top-left, top-right, bottom-left or
bottom-right corner. It combines the predictions from each corner and
center pair to generate a bounding box. Finally, it merges the four
bounding boxes to give a bounding box. CornerNet is very different from
PLN. First, CornerNet groups the corners by predicting embedding vectors,
while PLN groups the corner and center by predicting pixel locations.
Second, CornerNet uses corner pooling to better localize the corners.

Our approach is inspired by~\cite{newell2017associative} on Associative
Embedding in the context of multi-person pose estimation. Newell et al.
propose an approach that detects and groups human joints in a single
network. In their approach each detected human joint has an embedding
vector. The joints are grouped based on the distances between their
embeddings. To the best of our knowledge, we are the first to formulate the
task of object detection as a task of detecting and grouping corners with
embeddings. Another novelty of ours is the corner pooling layers that help
better localize the corners. We also significantly modify the hourglass
architecture and add our novel variant of focal loss~\citep{lin2017focal}
to help better train the network. 

\section{CornerNet}
\label{sec:cornernet}
\subsection{Overview}
In CornerNet, we detect an object as a pair of keypoints---the top-left
corner and bottom-right corner of the bounding box. A convolutional network
predicts two sets of heatmaps to represent the locations of corners of
different object categories, one set for the top-left corners and the other
for the bottom-right corners.  The network also predicts an embedding
vector for each detected corner~\citep{newell2017associative} such that the
distance between the embeddings of two corners from the same object is
small. To produce tighter bounding boxes, the network also predicts offsets
to slightly adjust the locations of the corners.  With the predicted
heatmaps, embeddings and offsets, we apply a simple post-processing
algorithm to obtain the final bounding boxes.

\begin{figure*} 
    \centering
    \resizebox{\textwidth}{!}{\includegraphics{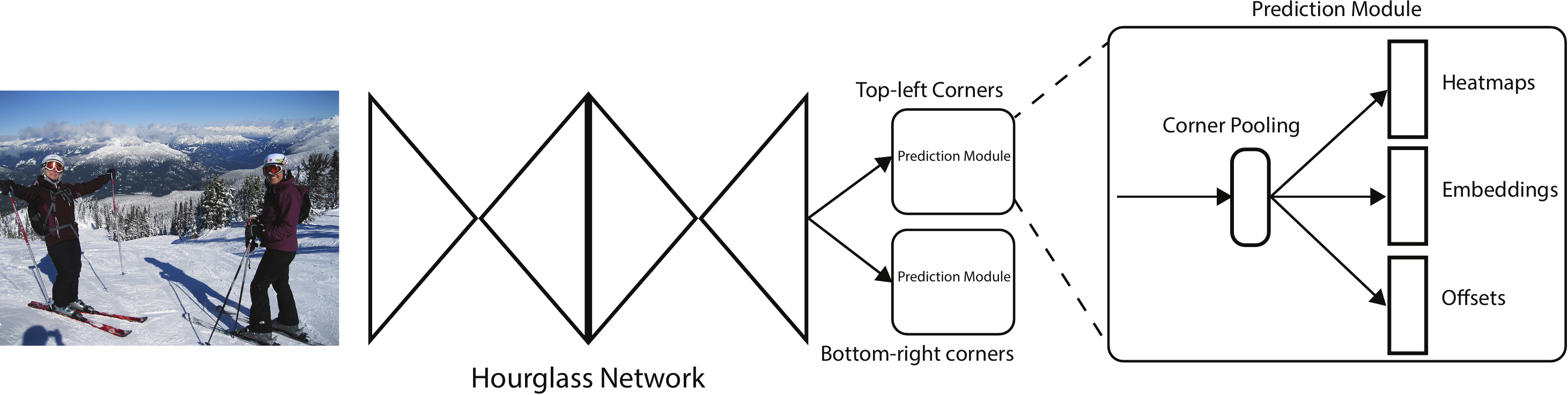}}
    \caption{Overview of CornerNet. The backbone network is followed by
    two prediction modules, one for the top-left corners and the other
    for the bottom-right corners. Using the predictions from both
    modules, we locate and group the corners. } 
    \label{fig:overview}
\end{figure*}

Fig.~\ref{fig:overview} provides an overview of CornerNet.  We use the
hourglass network~\citep{newell2016stacked} as the backbone network of
CornerNet. The hourglass network is followed by two prediction modules. One
module is for the top-left corners, while the other one is for the
bottom-right corners.  Each module has its own corner pooling module to
pool features from the hourglass network before predicting the heatmaps,
embeddings and offsets.  Unlike many other object detectors, we do not use
features from different scales to detect objects of different sizes. We
only apply both modules to the output of the hourglass network.  

\subsection{Detecting Corners}
\label{sec:detecting_corners}
We predict two sets of heatmaps, one for top-left corners and one for
bottom-right corners. Each set of heatmaps has $C$ channels, where $C$ is
the number of categories, and is of size $H \times W$. There is no
background channel. Each channel is a binary mask indicating the locations
of the corners for a class.

\begin{figure}
    \centering
    \resizebox{\columnwidth}{!}{\includegraphics{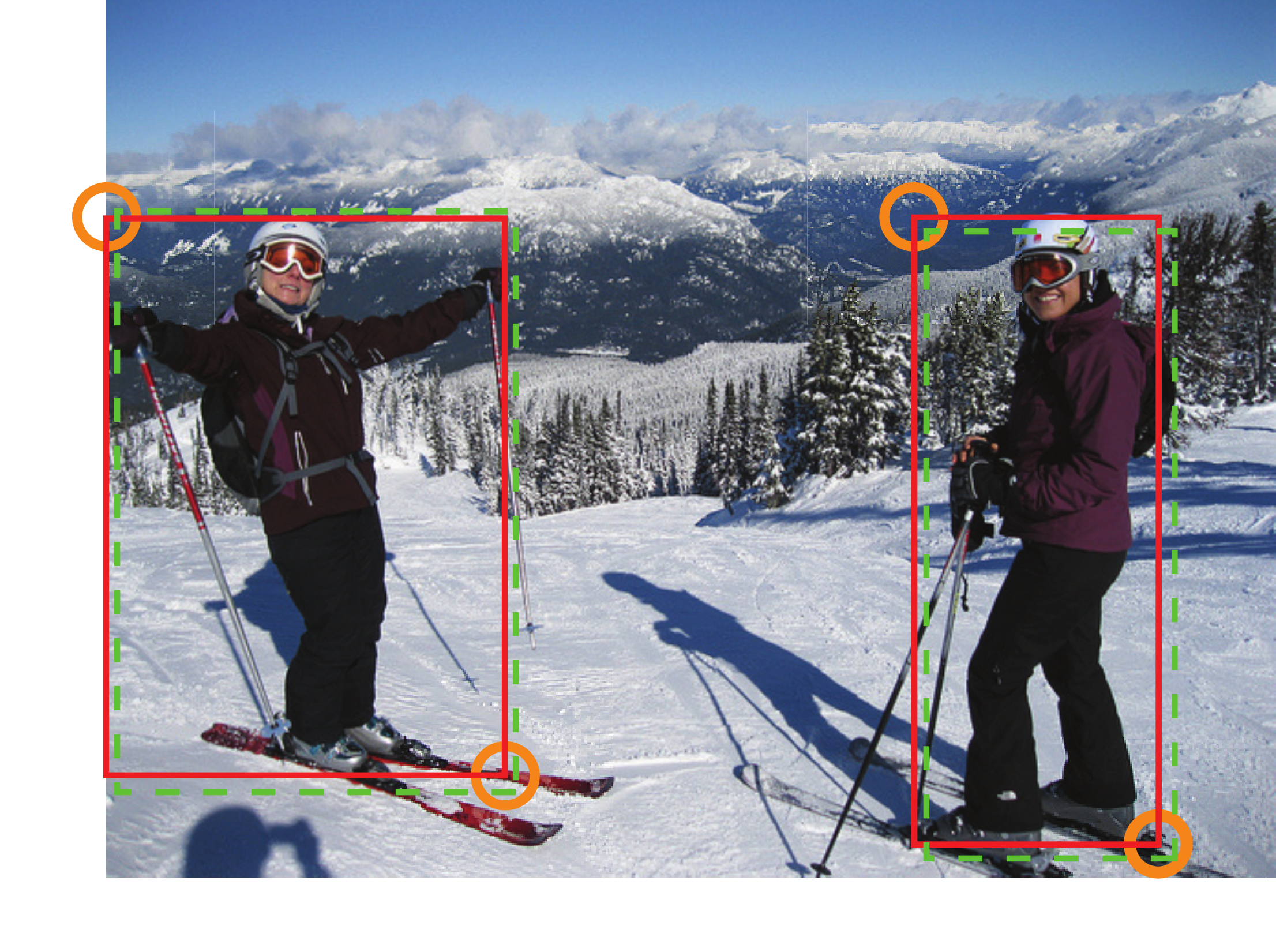}}
    \caption{``Ground-truth'' heatmaps for training. Boxes \textit{(green dotted rectangles)}
    whose corners are within the radii of the positive locations
    \textit{(orange circles)} still have large overlaps with the
    ground-truth annotations \textit{(red solid rectangles)}.}
    \label{fig:gt_heatmaps}
\end{figure}

For each corner, there is one ground-truth positive location, and all other
locations are negative. During training, instead of equally penalizing
negative locations, we reduce the penalty given to negative locations
within a radius of the positive location. This is because a pair of false
corner detections, if they are close to their respective ground truth
locations, can still produce a box that sufficiently overlaps the
ground-truth box (Fig.~\ref{fig:gt_heatmaps}).  We determine the radius by
the size of an object by ensuring that a pair of points within the radius
would generate a bounding box with at least $t$ IoU with the ground-truth
annotation (we set $t$ to $0.3$ in all experiments). Given the radius, the
amount of penalty reduction is given by an unnormalized 2D Gaussian,
$e^{-\frac{x^2 + y^2}{2 \sigma^2}}$, whose center is at the positive
location and whose $\sigma$ is 1/3 of the radius. 

Let $p_{cij}$ be the score at location $(i,j)$ for class $c$ in the
predicted heatmaps, and let $y_{cij}$ be the ``ground-truth'' heatmap
augmented with the unnormalized Gaussians. We design a variant of focal
loss~\citep{lin2017focal}: 
\begin{equation}
    \resizebox{0.90\columnwidth}{!}{$
    L_{det} = \frac{-1}{N} \sum\limits_{c=1}^{C} \sum\limits_{i=1}^{H}
    \sum\limits_{j=1}^{W}
    \left\{\begin{matrix}
        \left(1 - p_{cij}\right)^{\alpha} 
        \log\left(p_{cij}\right) & 
        \text{if}\ y_{cij} = 1\\ 
        \left(1 - y_{cij}\right)^{\beta} 
        \left(p_{cij}\right)^{\alpha}
        \log\left(1 - p_{cij}\right) & 
        \text{otherwise}
    \end{matrix}\right.
    $}
    \label{eq:det_loss}
\end{equation}
where $N$ is the number of objects in an image, and $\alpha$ and $\beta$
are the hyper-parameters which control the contribution of each point (we
set $\alpha$ to 2 and $\beta$ to 4 in all experiments). With the Gaussian
bumps encoded in $y_{cij}$, the $(1-y_{cij})$ term reduces the penalty
around the ground truth locations. 

Many networks~\citep{he2016deep,newell2016stacked} involve downsampling
layers to gather global information and to reduce memory usage. When they
are applied to an image fully convolutionally, the size of the output is
usually smaller than the image. Hence, a location $(x, y)$ in the image is
mapped to the location $\left(\lfloor\frac{x}{n}\rfloor,
\lfloor\frac{y}{n}\rfloor\right)$ in the heatmaps, where $n$ is the
downsampling factor. When we remap the locations from the heatmaps to the
input image, some precision may be lost, which can greatly affect the IoU
of small bounding boxes with their ground truths. To address this issue we
predict location offsets to slightly adjust the corner locations before
remapping them to the input resolution.
\begin{equation}
    \bm{o}_{k} = 
    \left( \frac{x_{k}}{n} - \left\lfloor\frac{x_{k}}{n}\right\rfloor, 
    \frac{y_{k}}{n} - \left\lfloor\frac{y_{k}}{n}\right\rfloor \right)
    \label{eq:delta_ij}
\end{equation}
where $\bm{o}_{k}$ is the offset, $x_{k}$ and $y_{k}$ are the x and y
coordinate for corner $k$. In particular, we predict one set of offsets
shared by the top-left corners of all categories, and another set shared by
the bottom-right corners. For training, we apply the smooth L1
Loss~\citep{girshick2015fast} at ground-truth corner locations: 
\begin{equation}
    L_{\mathit{off}} = \frac{1}{N} \sum^{N}_{k=1}
    \text{SmoothL1Loss}\left(\bm{o}_{k}, \bm{\hat{o}}_{k}\right) 
    \label{eq:offset_loss}
\end{equation}

\subsection{Grouping Corners}
Multiple objects may appear in an image, and thus multiple top-left and
bottom-right corners may be detected. We need to determine if a pair of the
top-left corner and bottom-right corner is from the same bounding box. Our
approach is inspired by the Associative Embedding method proposed
by~\cite{newell2017associative} for the task of multi-person pose
estimation. Newell et al. detect all human joints and generate an embedding
for each detected joint. They group the joints based on the distances
between the embeddings.

The idea of associative embedding is also applicable to our task. The
network predicts an embedding vector for each detected corner such that if
a top-left corner and a bottom-right corner belong to the same bounding
box, the distance between their embeddings should be small. We can then
group the corners based on the distances between the embeddings of the
top-left and bottom-right corners. The actual values of the embeddings are
unimportant. Only the distances between the embeddings are used to group
the corners.

We follow~\cite{newell2017associative} and use embeddings of 1 dimension.
Let $e_{t_{k}}$ be the embedding for the top-left corner of object $k$ and
$e_{b_{k}}$ for the bottom-right corner.  As in~\cite{newell2017pixels},
we use the ``pull'' loss to train the network to group the corners and the
``push'' loss to separate the corners:
\begin{equation}
    L_{pull} = \frac{1}{N} \sum^{N}_{k=1}
    \left[ \left( e_{t_{k}} - e_{k} \right)^2 +
    \left( e_{b_{k}} - e_{k}\right)^2 \right], 
    \label{eq:pull_loss}
\end{equation}
\begin{equation}
    L_{push} = \frac{1}{N(N - 1)} \sum^{N}_{k=1} 
    \sum^{N}\limits_{\substack{j=1 \\ j \neq k}}
    \max \left(0, \Delta - \left| e_{k} - e_{j} \right| \right), 
    \label{eq:push_loss}
\end{equation}
where $e_{k}$ is the average of $e_{t_{k}}$ and $e_{b_{k}}$ and we set
$\Delta$ to be $1$ in all our experiments. Similar to the offset loss, we
only apply the losses at the ground-truth corner location. 

\subsection{Corner Pooling}
\label{sec:pooling}
As shown in Fig.~\ref{fig:intro_local_evidence}, there is often no local
visual evidence for the presence of corners. To determine if a pixel is a
top-left corner, we need to look horizontally towards the right for the
topmost boundary of an object and vertically towards the bottom for the
leftmost boundary. We thus propose \textit{corner pooling} to better
localize the corners by encoding explicit prior knowledge. 

\begin{figure*}
    \centering
    \resizebox{0.6\textwidth}{!}{\includegraphics{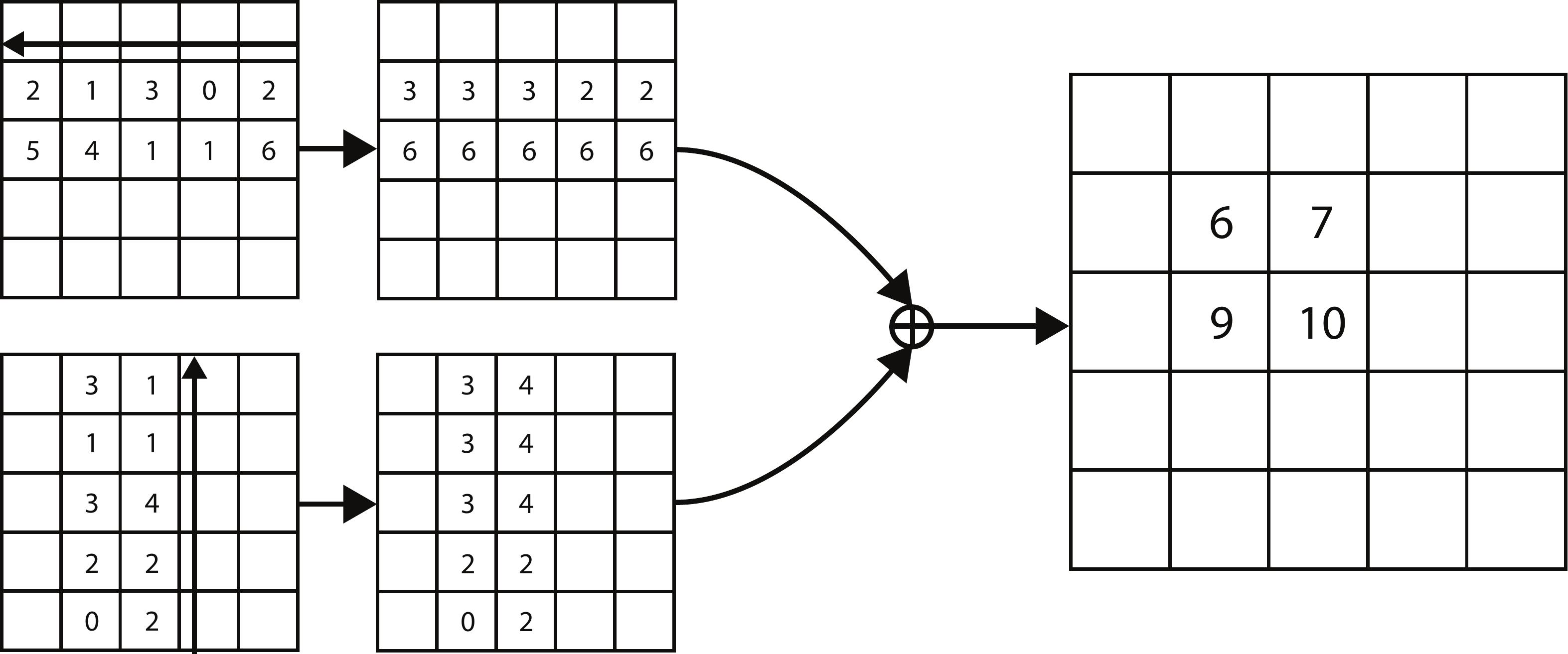}}
    \caption{The top-left corner pooling layer can be implemented very
    efficiently. We scan from right to left for the horizontal max-pooling
    and from bottom to top for the vertical max-pooling. We then add two
    max-pooled feature maps.}
    \label{fig:pooling}
\end{figure*}

Suppose we want to determine if a pixel at location $(i, j)$ is a top-left
corner. Let $f_{t}$ and $f_{l}$ be the feature maps that are the inputs to
the top-left corner pooling layer, and let $f_{t_{ij}}$ and $f_{l_{ij}}$ be
the vectors at location $(i, j)$ in $f_{t}$ and $f_{l}$ respectively.  With
$H\times W$ feature maps, the corner pooling layer first max-pools all
feature vectors between $(i, j)$ and $(i, H)$ in $f_{t}$ to a feature
vector $t_{ij}$, and max-pools all feature vectors between $(i, j)$ and
$(W, j)$ in $f_{l}$ to a feature vector $l_{ij}$. Finally, it adds $t_{ij}$
and $l_{ij}$ together. This computation can be expressed by the following
equations:
\begin{equation}
    t_{ij} = 
    \left\{\begin{matrix}
        \max \left(f_{t_{ij}}, t_{(i+1)j}\right) & \text{if}\ i < H \\
        f_{t_{Hj}} & \text{otherwise}
    \end{matrix}\right.
    \label{eq:top_pool}
\end{equation}

\begin{equation}
    l_{ij} = 
    \left\{\begin{matrix}
        \max \left(f_{l_{ij}}, l_{i(j+1)}\right) & \text{if}\ j < W \\
        f_{l_{iW}} & \text{otherwise}
    \end{matrix}\right.
    \label{eq:left_pool}
\end{equation}
where we apply an elementwise $\max$ operation. Both $t_{ij}$ and $l_{ij}$
can be computed efficiently by dynamic programming as shown
Fig.~\ref{fig:pooling}.

We define bottom-right corner pooling layer in a similar way. It max-pools
all feature vectors between $(0, j)$ and $(i, j)$, and all feature vectors
between $(i, 0)$ and $(i, j)$ before adding the pooled results. The corner
pooling layers are used in the prediction modules to predict heatmaps,
embeddings and offsets.

\begin{figure*}
    \centering
    \resizebox{0.8\textwidth}{!}{\includegraphics{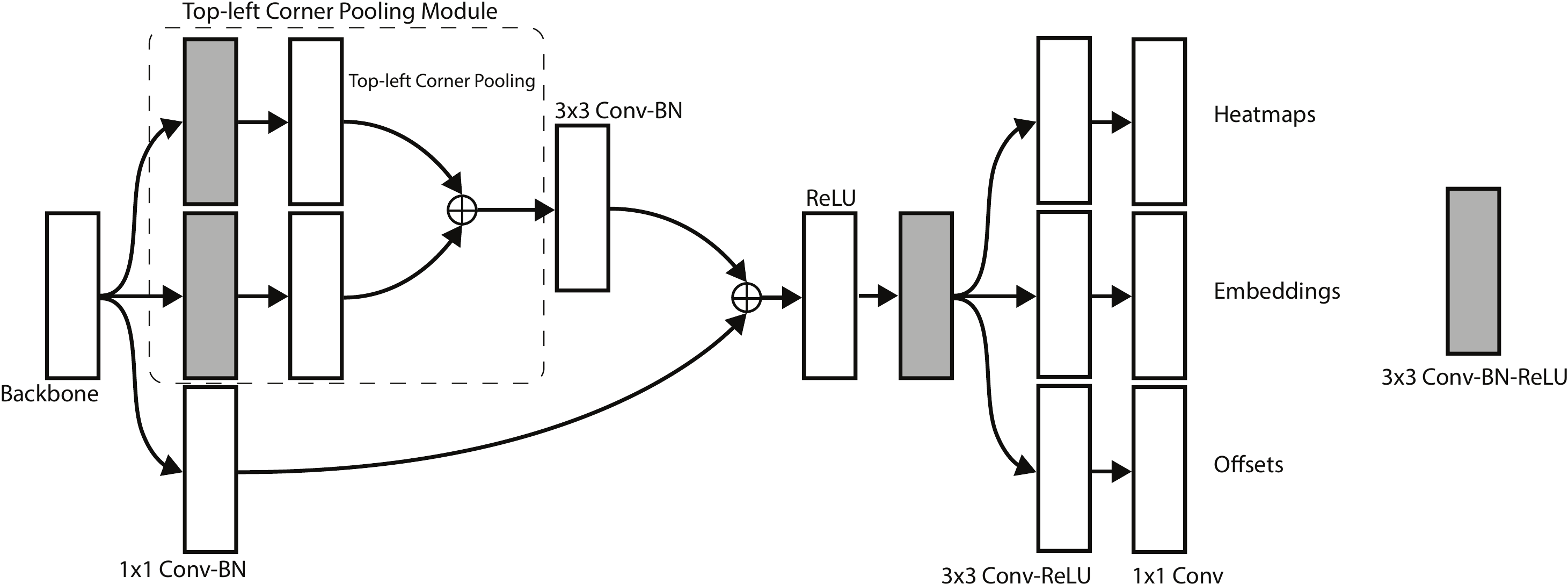}}
    \caption{The prediction module starts with a modified residual block,
    in which we replace the first convolution module with our corner
    pooling module. The modified residual block is then followed by a
    convolution module. We have multiple branches for predicting the 
    heatmaps, embeddings and offsets.}
    \label{eq:prediction_module}
\end{figure*}
The architecture of the prediction module is shown in
Fig.~\ref{eq:prediction_module}. The first part of the module is a modified
version of the residual block~\citep{he2016deep}. In this modified residual
block, we replace the first $3 \times 3$ convolution module with a corner
pooling module, which first processes the features from the backbone
network by two $3 \times 3$ convolution modules~\footnote{Unless otherwise
specified, our convolution module consists of a convolution layer, a BN
layer~\citep{ioffe2015batch} and a ReLU layer} with 128 channels and then
applies a corner pooling layer. Following the design of a residual block,
we then feed the pooled features into a $3 \times 3$ Conv-BN layer with 256
channels and add back the projection shortcut. The modified residual block
is followed by a $3 \times 3$ convolution module with 256 channels, and 3
Conv-ReLU-Conv layers to produce the heatmaps, embeddings and offsets.

\subsection{Hourglass Network}
CornerNet uses the hourglass network~\citep{newell2016stacked} as its
backbone network. The hourglass network was first introduced for the human
pose estimation task. It is a fully convolutional neural network that
consists of one or more hourglass modules. An hourglass module first
downsamples the input features by a series of convolution and max pooling
layers. It then upsamples the features back to the original resolution by a
series of upsampling and convolution layers. Since details are lost in the
max pooling layers, skip layers are added to bring back the details to the
upsampled features. The hourglass module captures both global and local
features in a single unified structure. When multiple hourglass modules are
stacked in the network, the hourglass modules can reprocess the features to
capture higher-level of information. These properties make the hourglass
network an ideal choice for object detection as well. In fact, many current
detectors~\citep{shrivastava2016beyond,fu2017dssd,lin2016feature,kong2017ron}
already adopted networks similar to the hourglass network.

Our hourglass network consists of two hourglasses, and we make some
modifications to the architecture of the hourglass module. Instead of using
max pooling, we simply use stride 2 to reduce feature resolution. We reduce
feature resolutions 5 times and increase the number of feature channels
along the way ($256, 384, 384, 384, 512$). When we upsample the features,
we apply 2 residual modules followed by a nearest neighbor upsampling.
Every skip connection also consists of 2 residual modules. There are 4
residual modules with 512 channels in the middle of an hourglass module.
Before the hourglass modules, we reduce the image resolution by 4 times
using a $7 \times 7$ convolution module with stride 2 and 128 channels
followed by a residual block~\citep{he2016deep} with stride 2 and 256
channels.

Following~\citep{newell2016stacked}, we also add intermediate supervision
in training. However, we do not add back the intermediate predictions to
the network as we find that this hurts the performance of the network. We
apply a $1 \times 1$ Conv-BN module to both the input and output of the
first hourglass module. We then merge them by element-wise addition
followed by a ReLU and a residual block with 256 channels, which is then
used as the input to the second hourglass module.  The depth of the
hourglass network is 104. Unlike many other state-of-the-art detectors, we
only use the features from the last layer of the whole network to make
predictions. 

\begin{table*}
    \begin{center}
    \caption{Ablation on corner pooling on MS COCO validation.}
    \label{tab:pooling}
    \begin{tabularx}{0.7\textwidth}{|l|YYY|YYY|}
        \hline
        & AP & $\text{AP}^{50}$ & $\text{AP}^{75}$ & $\text{AP}^{s}$ &
        $\text{AP}^{m}$ & $\text{AP}^{l}$ \\ \hline \hline
        w/o corner pooling & 36.5 & 52.0 & 38.8 & 17.5 & 38.9 & 49.4 \\
        \hline
        w/ corner pooling  & 38.4 & 53.8 & 40.9 & 18.6 & 40.5 & 51.8 \\
        \hline \hline
        improvement & +2.0 & +2.1 & +2.1 & +1.1 & +2.4 & +3.6 \\ \hline
    \end{tabularx}
    \end{center}
\end{table*}

\begin{table*}
    \begin{center}
    \caption{Reducing the penalty given to the negative locations near
    positive locations helps significantly improve the performance of
    the network}
    \label{tab:radii}
    \begin{tabularx}{0.7\textwidth}{|l|YYY|YYY|}
        \hline
        & AP & $\text{AP}^{50}$ & $\text{AP}^{75}$ & $\text{AP}^{s}$ &
        $\text{AP}^{m}$ & $\text{AP}^{l}$ \\ \hline \hline
        w/o reducing penalty & 32.9 & 49.1 & 34.8 & 19.0 & 37.0 & 40.7 \\
        \hline
        fixed radius & 35.6 & 52.5 & 37.7 & 18.7 & 38.5 & 46.0 \\ \hline
        object-dependent radius & 38.4 & 53.8 & 40.9 & 18.6 & 40.5 & 51.8 \\
        \hline
    \end{tabularx}
    \end{center}
\end{table*}

\begin{table*}
    \begin{center}
        \caption{Corner pooling consistently improves the network
        performance on detecting corners in different image quadrants,
        showing that corner pooling is effective and stable over both
        small and large areas.}
        \label{tab:stability}
        \begin{tabularx}{0.7\textwidth}{|l|Y|Y|Y|} \hline
            & mAP w/o pooling & mAP w/ pooling & improvement \\ \hline
            \hline
            \multicolumn{4}{|l|}{\textbf{Top-Left Corners}} \\ \hline
            Top-Left Quad. & 66.1 & 69.2 & +3.1 \\ \hline
            Bottom-Right Quad. & 60.8 & 63.5 & +2.7 \\ \hline \hline
            \multicolumn{4}{|l|}{\textbf{Bottom-Right Corners}} \\ \hline
            Top-Left Quad. & 53.4 & 56.2 & +2.8 \\ \hline
            Bottom-Right Quad. & 65.0 & 67.6 & +2.6 \\ \hline
        \end{tabularx}
    \end{center}
\end{table*}

\section{Experiments}
\subsection{Training Details}
\label{sec:training}
We implement CornerNet in PyTorch~\citep{paszke2017automatic}.  The network
is randomly initialized under the default setting of PyTorch with no
pretraining on any external dataset. As we apply focal loss, we
follow~\citep{lin2017focal} to set the biases in the convolution layers
that predict the corner heatmaps.  During training, we set the input
resolution of the network to $511 \times 511$, which leads to an output
resolution of $128 \times 128$. To reduce overfitting, we adopt standard
data augmentation techniques including random horizontal flipping, random
scaling, random cropping and random color jittering, which includes
adjusting the brightness, saturation and contrast of an image.  Finally, we
apply PCA~\citep{krizhevsky2012imagenet} to the input image.

We use Adam~\citep{kingma2014adam} to optimize the full training loss: 
\begin{equation}
    L = L_{det} + \alpha L_{pull} + \beta L_{push} + \gamma L_{\mathit{off}}
    \label{eq:training_loss}
\end{equation}
where $\alpha$, $\beta$ and $\gamma$ are the weights for the pull, push and
offset loss respectively. We set both $\alpha$ and $\beta$ to 0.1 and
$\gamma$ to 1. We find that 1 or larger values of $\alpha$ and $\beta$ lead
to poor performance. We use a batch size of 49 and train the network on 10
Titan X (PASCAL) GPUs (4 images on the master GPU, 5 images per GPU for the
rest of the GPUs). To conserve GPU resources, in our ablation experiments,
we train the networks for 250k iterations with a learning rate of $2.5
\times 10^{-4}$. When we compare our results with other detectors, we train
the networks for an extra 250k iterations and reduce the learning rate to
$2.5 \times 10^{-5}$ for the last 50k iterations.

\begin{figure*}
    \centering
    \resizebox{\textwidth}{!}{\includegraphics{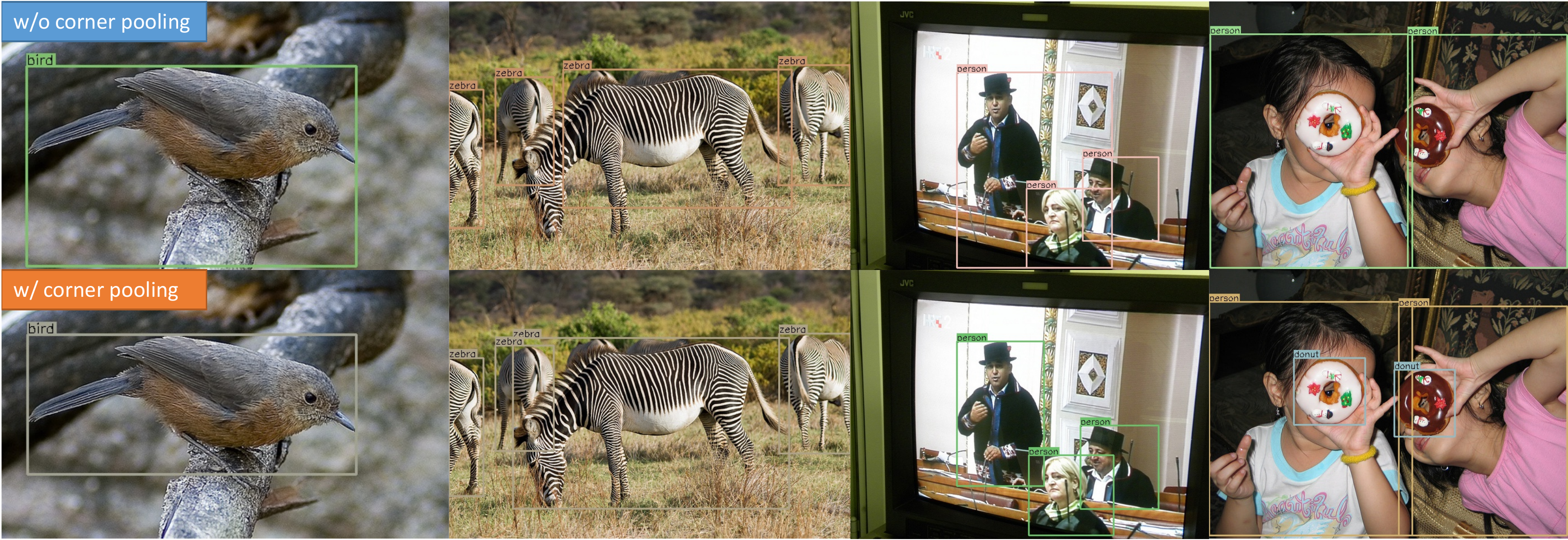}}
    \caption{Qualitative examples showing corner pooling helps better
    localize the corners.}
    \label{fig:pooling}
\end{figure*}

\subsection{Testing Details}
\label{sec:testing}
During testing, we use a simple post-processing algorithm to generate
bounding boxes from the heatmaps, embeddings and offsets. We first apply
non-maximal suppression (NMS) by using a $3 \times 3$ max pooling layer on
the corner heatmaps. Then we pick the top 100 top-left and top 100
bottom-right corners from the heatmaps. The corner locations are adjusted
by the corresponding offsets. We calculate the L1 distances between the
embeddings of the top-left and bottom-right corners. Pairs that have
distances greater than 0.5 or contain corners from different categories are
rejected. The average scores of the top-left and bottom-right corners are
used as the detection scores.

Instead of resizing an image to a fixed size, we maintain the original
resolution of the image and pad it with zeros before feeding it to
CornerNet. Both the original and flipped images are used for testing. We
combine the detections from the original and flipped images, and apply
soft-nms~\citep{bodla2017soft} to suppress redundant detections. Only the
top 100 detections are reported. The average inference time is 244ms per
image on a Titan X (PASCAL) GPU.

\subsection{MS COCO}
We evaluate CornerNet on the very challenging MS COCO
dataset~\citep{lin2014microsoft}.  MS COCO contains 80k images for
training, 40k for validation and 20k for testing. All images in the
training set and 35k images in the validation set are used for training.
The remaining 5k images in validation set are used for hyper-parameter
searching and ablation study. All results on the test set are submitted to
an external server for evaluation. To provide fair comparisons with other
detectors, we report our main results on the test-dev set. MS COCO uses
average precisions (APs) at different IoUs and APs for different object
sizes as the main evaluation metrics.

\begin{table*}
    \begin{center}
    \caption{The hourglass network is crucial to the performance of
    CornerNet.}
    \label{tab:hourglass}
    \begin{tabularx}{0.7\textwidth}{|l|YYY|YYY|}
        \hline
        & AP & $\text{AP}^{50}$ & $\text{AP}^{75}$ & $\text{AP}^{s}$ &
        $\text{AP}^{m}$ & $\text{AP}^{l}$ \\ \hline \hline
        FPN (w/ ResNet-101) + Corners & 30.2 & 44.1 & 32.0 & 13.3 & 33.3 & 42.7 \\ \hline
        Hourglass + Anchors & 32.9 & 53.1 & 35.6 & 16.5 & 38.5 & 45.0 \\ \hline
        Hourglass + Corners & 38.4 & 53.8 & 40.9 & 18.6 & 40.5 & 51.8 \\ \hline
    \end{tabularx}
    \end{center}
\end{table*}

\subsection{Ablation Study}
\subsubsection{Corner Pooling}
Corner pooling is a key component of CornerNet. To understand its
contribution to performance, we train another network without corner
pooling but with the same number of parameters. 

Tab.~\ref{tab:pooling} shows that adding corner pooling gives significant
improvement: 2.0\% on AP, 2.1\% on $\text{AP}^{50}$ and 2.1\% on
$\text{AP}^{75}$. We also see that corner pooling is especially helpful for
medium and large objects, improving their APs by 2.4\% and 3.6\%
respectively. This is expected because the topmost, bottommost, leftmost,
rightmost boundaries of medium and large objects are likely to be further
away from the corner locations. Fig.~\ref{fig:pooling} shows four
qualitative examples with and without corner pooling.

\begin{table*}
    \begin{center}
    \caption{CornerNet performs much better at high IoUs than other
    state-of-the-art detectors.}
    \label{tab:ious}
    \begin{tabularx}{0.8\textwidth}{|l|Y|YYYYY|}
        \hline
        & AP & $\text{AP}^{50}$ & $\text{AP}^{60}$ & $\text{AP}^{70}$ &
        $\text{AP}^{80}$ & $\text{AP}^{90}$ \\ \hline \hline
        RetinaNet~\citep{lin2017focal} & 39.8 & 59.5 & 55.6 & 48.2 & 36.4 &
        15.1 \\ \hline
        Cascade R-CNN~\citep{cai2017cascade} & 38.9 & 57.8 & 53.4 & 46.9 &
        35.8 & 15.8 \\ \hline
        Cascade R-CNN + IoU Net~\citep{jiang2018acquisition} & 41.4 & 59.3 &
        55.3 & 49.6 & 39.4 & 19.5 \\ \hline
        CornerNet & 40.6 & 56.1 & 52.0 & 46.8 & 38.8 & 23.4 \\ \hline
    \end{tabularx}
    \end{center}
\end{table*}

\begin{table*}
    \begin{center}
    \caption{Error analysis. We replace the predicted heatmaps and offsets
        with the ground-truth values. Using the ground-truth heatmaps alone
        improves the AP from 38.4\% to 73.1\%, suggesting that the main
        bottleneck of CornerNet is detecting corners. }
    \label{tab:predicted_gts}
    \begin{tabularx}{0.7\textwidth}{|l|YYY|YYY|}
        \hline
        & AP & $\text{AP}^{50}$ & $\text{AP}^{75}$ & $\text{AP}^{s}$ &
        $\text{AP}^{m}$ & $\text{AP}^{l}$ \\ \hline \hline
        & 38.4 & 53.8 & 40.9 & 18.6 & 40.5 & 51.8 \\ \hline
        w/ gt heatmaps & 73.1 & 87.7 & 78.4 & 60.9 & 81.2 & 81.8 \\ \hline
        w/ gt heatmaps + offsets & 86.1 & 88.9 & 85.5 & 84.8 & 87.2 & 82.0
        \\ \hline
    \end{tabularx}
    \end{center}
\end{table*}

\begin{figure*}
    \centering
    \resizebox{\textwidth}{!}{\includegraphics{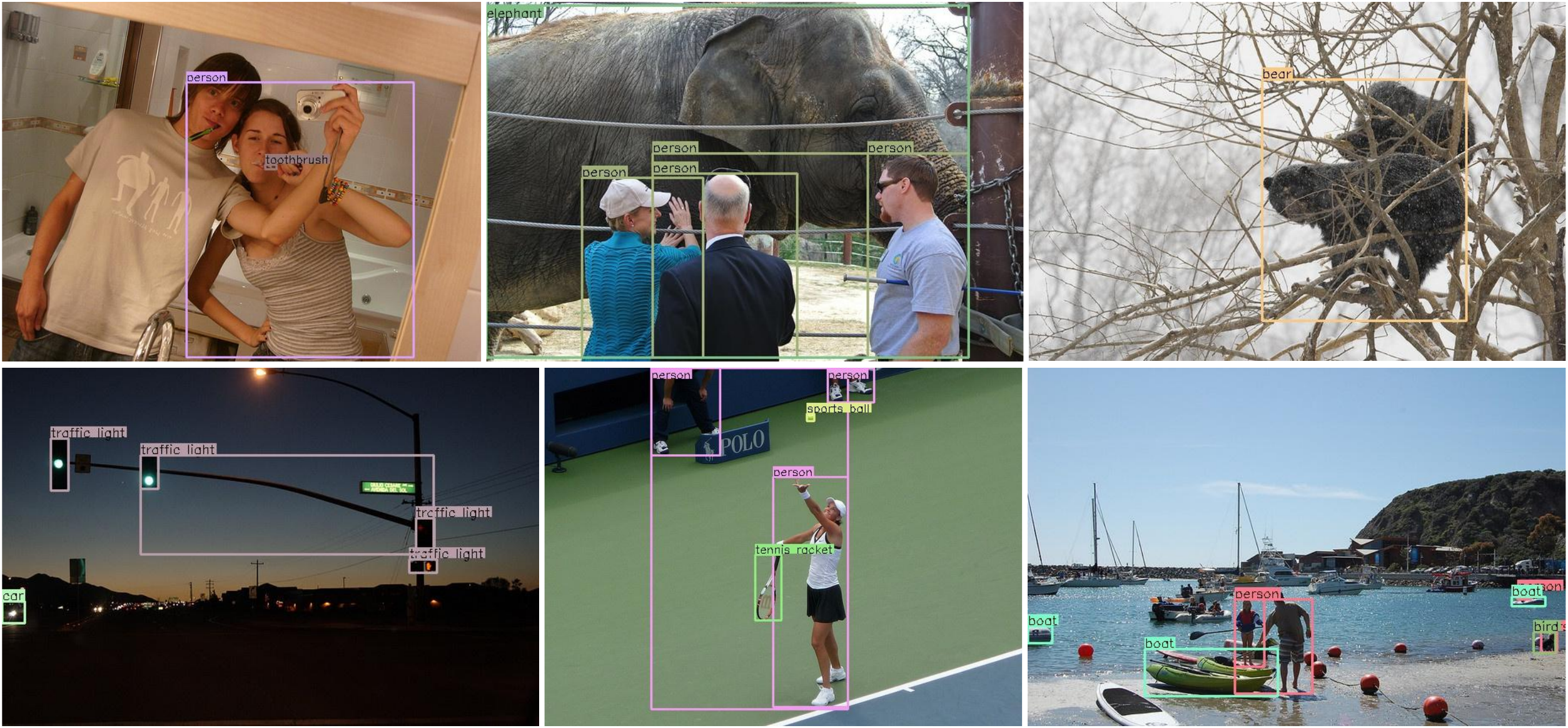}}
    \caption{Qualitative example showing errors in predicting corners and
    embeddings. The first row shows images where CornerNet mistakenly
    combines boundary evidence from different objects. The second row shows
    images where CornerNet predicts similar embeddings for corners from
    different objects.}
    \label{fig:error}
\end{figure*}

\begin{table*}
    \begin{center}
    \caption{CornerNet versus others on MS COCO test-dev.
    CornerNet outperforms all one-stage detectors and achieves results
    competitive to two-stage detectors}
    \label{tab:test}
    \resizebox{\textwidth}{!}{
        \begin{tabular}{|l|l|ccc|ccc|ccc|ccc|} \hline
        Method & Backbone & AP & $\text{AP}^{50}$ & $\text{AP}^{75}$ & 
        $\text{AP}^{s}$ & $\text{AP}^{m}$  & $\text{AP}^{l}$   &
        $\text{AR}^{1}$ & $\text{AR}^{10}$ & $\text{AR}^{100}$ &
        $\text{AR}^{s}$ & $\text{AR}^{m}$  & $\text{AR}^{l}$   \\
        \hline
        \hline
        \multicolumn{14}{|l|}{\textbf{Two-stage detectors}} \\ \hline
        DeNet~\citep{tychsen2017denet} & ResNet-101 & 33.8 & 53.4 & 36.1 &
        12.3 & 36.1 & 50.8 & 29.6 & 42.6 & 43.5 & 19.2 & 46.9 & 64.3\\
        \hline 
        CoupleNet~\citep{zhu2017couplenet} & ResNet-101 & 34.4 & 54.8 & 37.2
        & 13.4 & 38.1 & 50.8 & 30.0 & 45.0 & 46.4 & 20.7 & 53.1 & 68.5 \\
        \hline 
        Faster R-CNN by G-RMI~\citep{huang2017speed} &
        Inception-ResNet-v2~\citep{szegedy2017inception} & 34.7 & 55.5 &
        36.7 & 13.5 & 38.1 & 52.0 & - & - & - & - & - & -\\ \hline
        Faster R-CNN+++~\citep{he2016deep} & ResNet-101 & 34.9 & 55.7 & 37.4
        & 15.6 & 38.7 & 50.9 & - & - & - & - & - & - \\ \hline 
        Faster R-CNN w/ FPN~\citep{lin2016feature} & ResNet-101 & 36.2 &
        59.1 & 39.0 & 18.2 & 39.0 & 48.2 & - & - & - & - & - & - \\ \hline
        Faster R-CNN w/ TDM~\citep{shrivastava2016beyond} &
        Inception-ResNet-v2 & 36.8 & 57.7 & 39.2 & 16.2 & 39.8 & 52.1 &
        31.6 & 49.3 & 51.9 & 28.1 & 56.6 & 71.1 \\ \hline 
        D-FCN~\citep{dai2017deformable} & Aligned-Inception-ResNet & 37.5 &
        58.0 & - & 19.4 & 40.1 & 52.5 & - & - & - & - & - & -\\ \hline
        Regionlets~\citep{xu2017deep} & ResNet-101 & 39.3 & 59.8 & - & 21.7
        & 43.7 & 50.9 & - & - & - & - & - & - \\ \hline 
        Mask R-CNN~\citep{he2017mask} & ResNeXt-101 & 39.8 & 62.3 & 43.4 &
        22.1 & 43.2 & 51.2 & - & - & - & - & - & -\\ \hline 
        Soft-NMS~\citep{bodla2017soft} &
        Aligned-Inception-ResNet & 40.9 & 62.8 & - & 23.3 & 43.6 & 53.3 & -
        & - & - & - & - & -\\ \hline 
        LH R-CNN~\citep{li2017light} & ResNet-101 & 41.5 & - & - & 25.2 &
        45.3 & 53.1 & - & - & - & - & - & - \\ \hline 
        Fitness-NMS~\citep{tychsen2017improving} & ResNet-101
        & 41.8 & 60.9 & 44.9 & 21.5 & 45.0 & 57.5 & - & - & - & - & - & -\\
        \hline 
        Cascade R-CNN~\citep{cai2017cascade} & ResNet-101 & 42.8 & 62.1 &
        46.3 & 23.7 & 45.5 & 55.2 & - & - & - & - & - & -\\ \hline 
        D-RFCN + SNIP~\citep{singh2017analysis} & DPN-98~\citep{chen2017dual}
        & 45.7 & 67.3 & 51.1 & 29.3 & 48.8 & 57.1 & - & - & - & - & - & -
        \\ \hline \hline
        \multicolumn{14}{|l|}{\textbf{One-stage detectors}} \\ \hline
        YOLOv2~\citep{redmon2016yolo9000} & DarkNet-19 & 21.6 & 44.0 & 19.2
        & 5.0 & 22.4 & 35.5 & 20.7 & 31.6 & 33.3 & 9.8 & 36.5 & 54.4\\
        \hline 
        DSOD300~\citep{shen2017dsod} & DS/64-192-48-1 & 29.3 & 47.3 & 30.6
        & 9.4 & 31.5 & 47.0 & 27.3 & 40.7 & 43.0 & 16.7 & 47.1 & 65.0\\
        \hline 
        GRP-DSOD320~\citep{shen2017learning} & DS/64-192-48-1 & 30.0 & 47.9
        & 31.8 & 10.9 & 33.6 & 46.3 & 28.0 & 42.1 & 44.5 & 18.8 & 49.1 &
        65.0 \\ \hline 
        SSD513~\citep{liu2016ssd} & ResNet-101 & 31.2 & 50.4 & 33.3 & 10.2 &
        34.5 & 49.8 & 28.3 & 42.1 & 44.4 & 17.6 & 49.2 & 65.8\\ \hline
        DSSD513~\citep{fu2017dssd} & ResNet-101 & 33.2 & 53.3 & 35.2 & 13.0
        & 35.4 & 51.1 & 28.9 & 43.5 & 46.2 & 21.8 & 49.1 & 66.4\\ \hline
        RefineDet512 (single scale)~\citep{zhang2017single} & ResNet-101 &
        36.4 & 57.5 & 39.5 & 16.6 & 39.9 & 51.4 & - & - & - & - & - & -\\
        \hline 
        RetinaNet800~\citep{lin2017focal} & ResNet-101 & 39.1 & 59.1 & 42.3
        & 21.8 & 42.7 & 50.2 & - & - & - & - & - & -\\ \hline 
        RefineDet512 (multi scale)~\citep{zhang2017single} & ResNet-101 &
        41.8 & 62.9 & 45.7 & 25.6 & 45.1 & 54.1 & - & - & - & - & - & -\\
        \thickhline 
        CornerNet511 (single scale) & Hourglass-104 & 40.6 & 56.4 & 43.2 &
        19.1 & 42.8 & 54.3 & 35.3 & 54.7 & 59.4 & 37.4 & 62.4 & 77.2 \\
        \hline
        CornerNet511 (multi scale) & Hourglass-104 & 42.2 & 57.8 & 45.2 &
        20.7 & 44.8 & 56.6 & 36.6 & 55.9 & 60.3 & 39.5 & 63.2 & 77.3
        \\ \hline
    \end{tabular}}
    \end{center}
\end{table*}

\subsubsection{Stability of Corner Pooling over Larger Area}
Corner pooling pools over different sizes of area in different quadrants of
an image. For example, the top-left corner pooling pools over larger areas
both horizontally and vertically in the upper-left quadrant of an image,
compared to the lower-right quadrant. Therefore, the location of a corner
may affect the stability of the corner pooling.

We evaluate the performance of our network on detecting both the top-left
and bottom-right corners in different quadrants of an image. Detecting
corners can be seen as a binary classification task i.e. the ground-truth
location of a corner is positive, and any location outside of a small
radius of the corner is negative. We measure the performance using mAPs
over all categories on the MS COCO validation set.

Tab.~\ref{tab:stability} shows that without corner pooling, the top-left
corner mAPs of upper-left and lower-right quadrant are $66.1\%$ and
$60.8\%$ respectively. Top-left corner pooling improves the mAPs by $3.1\%$
(to $69.2\%$) and $2.7\%$ (to $63.5\%$) respectively. Similarly,
bottom-right corner pooling improves the bottom-right corner mAPs of
upper-left quadrant by $2.8\%$ (from $53.4\%$ to $56.2\%$), and lower-right
quadrant by $2.6\%$ (from $65.0\%$ to $67.6\%$).  Corner pooling gives
similar improvement to corners at different quadrants, show that corner
pooling is effective and stable over both small and large areas.

\subsubsection{Reducing Penalty to Negative Locations} We reduce the
penalty given to negative locations around a positive location, within a
radius determined by the size of the object
(Sec.~\ref{sec:detecting_corners}). To understand how this helps train
CornerNet, we train one network with no penalty reduction and another
network with a fixed radius of $2.5$. We compare them with CornerNet on the
validation set.

Tab.~\ref{tab:radii} shows that a fixed radius improves AP over the baseline
by 2.7\%, $\text{AP}^{m}$ by 1.5\% and $\text{AP}^{l}$ by 5.3\%.
Object-dependent radius further improves the AP by 2.8\%, $\text{AP}^{m}$ by
2.0\% and $\text{AP}^{l}$ by 5.8\%. In addition, we see
that the penalty reduction especially benefits medium and large objects.

\subsubsection{Hourglass Network}
CornerNet uses the hourglass network~\citep{newell2016stacked} as its
backbone network. Since the hourglass network is not commonly used in other
state-of-the-art detectors, we perform an experiment to study the
contribution of the hourglass network in CornerNet. We train a CornerNet in
which we replace the hourglass network with FPN (w/
ResNet-101)~\citep{lin2017focal}, which is more commonly used in
state-of-the-art object detectors. We only use the final output of FPN for
predictions. Meanwhile, we train an anchor box based detector which uses
the hourglass network as its backbone. Each hourglass module predicts
anchor boxes at multiple resolutions by using features at multiple scales
during upsampling stage. We follow the anchor box design in
RetinaNet~\citep{lin2017focal} and add intermediate supervisions during
training. In both experiments, we initialize the networks from scratch and
follow the same training procedure as we train CornerNet
(Sec.~\ref{sec:training}).

Tab.~\ref{tab:hourglass} shows that CornerNet with hourglass network
outperforms CornerNet with FPN by $8.2\%$ AP, and the anchor box based
detector with hourglass network by $5.5\%$ AP. The results suggest that
the choice of the backbone network is important and the hourglass network
is crucial to the performance of CornerNet.

\begin{figure*}
   \centering
   \resizebox{\textwidth}{!}{\includegraphics{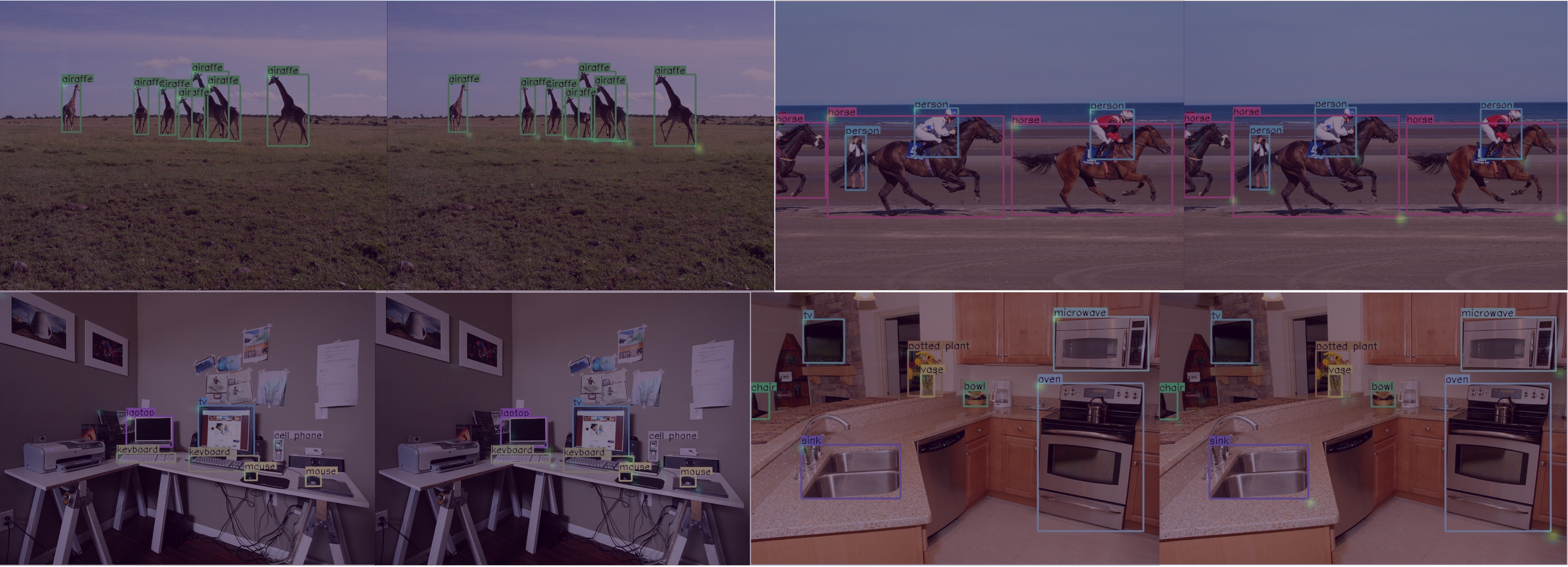}}
   \caption{Example bounding box predictions overlaid on predicted
   heatmaps of corners.}
   \label{fig:qualitative}
\end{figure*}

\subsubsection{Quality of the Bounding Boxes} 
A good detector should predict high quality bounding boxes that cover
objects tightly. To understand the quality of the bounding boxes predicted
by CornerNet, we evaluate the performance of CornerNet at multiple IoU
thresholds, and compare the results with other state-of-the-art detectors,
including RetinaNet~\citep{lin2017focal}, Cascade
R-CNN~\citep{cai2017cascade} and IoU-Net~\citep{jiang2018acquisition}.

Tab.~\ref{tab:ious} shows that CornerNet achieves a much higher AP at 0.9
IoU than other detectors, outperforming Cascade R-CNN + IoU-Net by
$3.9\%$, Cascade R-CNN by $7.6\%$ and RetinaNet~\footnote{We use the best
model publicly available on
\url{https://github.com/facebookresearch/Detectron/blob/master/MODEL_ZOO.md}}
by $7.3\%$. This suggests that CornerNet is able to generate bounding boxes
of higher quality compared to other state-of-the-art detectors. 

\begin{figure*}
    \centering
    \resizebox{\textwidth}{!}{\includegraphics{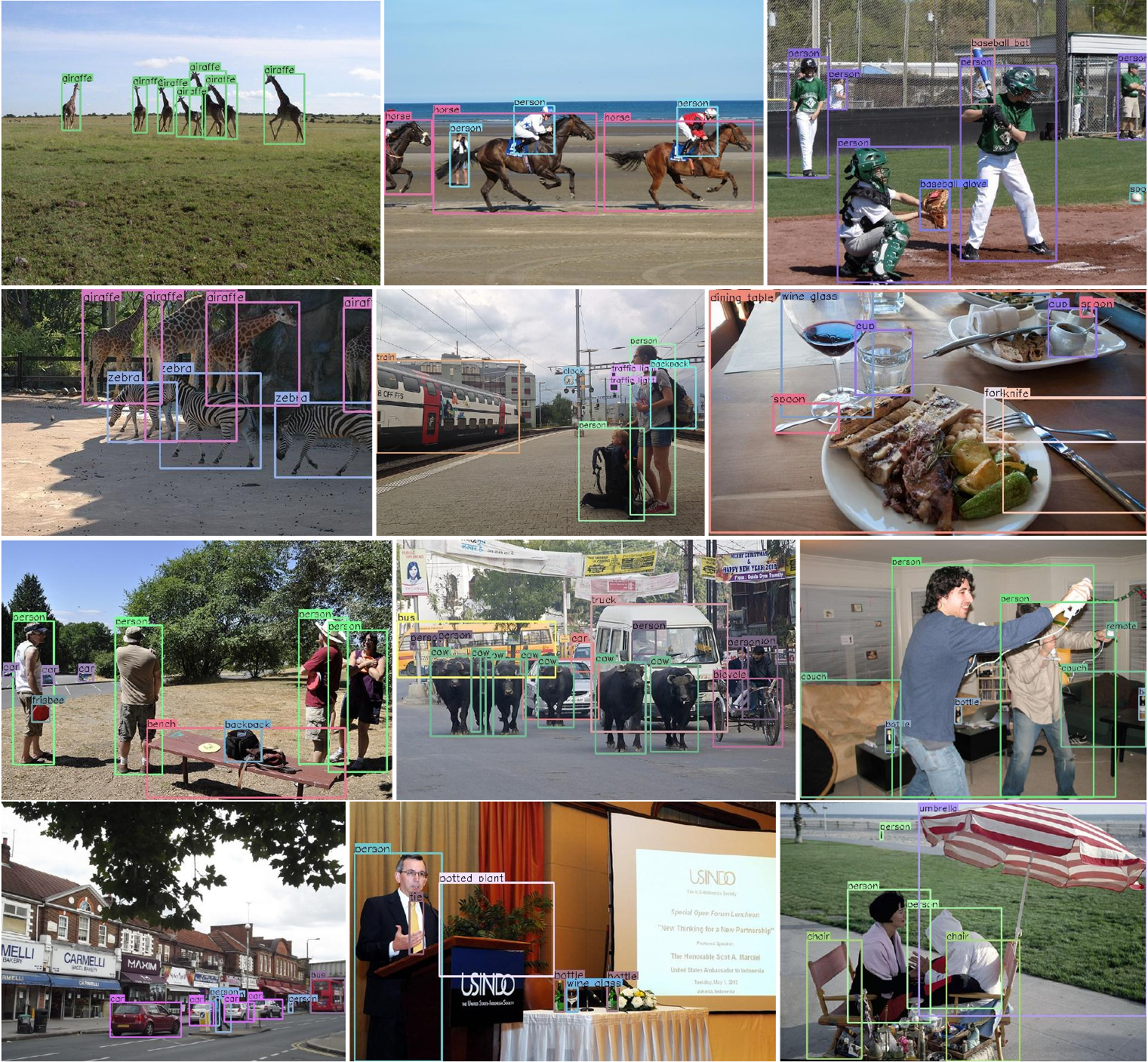}}
    \caption{Qualitative examples on MS COCO.}
    \label{fig:example}
\end{figure*}

\subsubsection{Error Analysis} CornerNet simultaneously outputs heatmaps,
offsets, and embeddings, all of which affect detection performance. An
object will be missed if either corner is missed; precise offsets are
needed to generate tight bounding boxes; incorrect embeddings will result
in many false bounding boxes. To understand how each part contributes to
the final error, we perform an error analysis by replacing the predicted
heatmaps and offsets with the ground-truth values and evaluting performance
on the validation set.  

Tab.~\ref{tab:predicted_gts} shows that using the ground-truth corner
heatmaps alone improves the AP from 38.4\% to 73.1\%. $\text{AP}^{s}$,
$\text{AP}^{m}$ and $\text{AP}^{l}$ also increase by 42.3\%, 40.7\% and
30.0\% respectively. If we replace the predicted offsets with the
ground-truth offsets, the AP further increases by 13.0\% to 86.1\%.  This
suggests that although there is still ample room for improvement in both
detecting and grouping corners, the main bottleneck is detecting corners.
Fig.~\ref{fig:error} shows some qualitative examples where the corner
locations or embeddings are incorrect.

\subsection{Comparisons with state-of-the-art detectors}
\label{sec:comparisons}
We compare CornerNet with other state-of-the-art detectors on MS COCO
test-dev (Tab.~\ref{tab:test}).  With multi-scale evaluation, CornerNet
achieves an AP of 42.2\%, the state of the art among existing one-stage
methods and competitive with two-stage methods. 

\section{Conclusion}
We have presented CornerNet, a new approach to object detection that
detects bounding boxes as pairs of corners. We evaluate CornerNet on MS
COCO and demonstrate competitive results. \\

\begin{acknowledgements}
    This work is partially supported by a grant from Toyota Research
    Institute and a DARPA grant FA8750-18-2-0019.  This article solely
    reflects the opinions and conclusions of its authors.
\end{acknowledgements}

\bibliographystyle{apalike}      
\bibliography{template}   

\end{document}